\pdfoutput=1
\documentclass[11pt]{article}

\usepackage[]{acl}
\usepackage{tikz-dependency}
\usepackage{times}
\usepackage{latexsym}
\usepackage{booktabs}
\usepackage{url}
\usepackage{threeparttable}
\usepackage[T1]{fontenc}
\usepackage{nert}

\usepackage{phaistos}
\usepackage[utf8]{inputenc}
\usepackage{listings}
\usepackage{microtype}

\usepackage{inconsolata}

\usepackage{graphicx}

\title{UD-English-CHILDES: A Collected Resource of Gold and Silver Universal~Dependencies Trees for Child Language Interactions}

\author{
Xiulin Yang\textsuperscript{\tiny{\PHplaneTree}} \quad 
Zhuoxuan Ju\textsuperscript{\tiny{\PHplaneTree}} \quad 
Lanni Bu\textsuperscript{\tiny{\PHplaneTree}} \quad 
Zoey Liu\textsuperscript{\tiny{\PHrosette}} \quad 
Nathan Schneider\textsuperscript{\tiny{\PHplaneTree}} \\
\textsuperscript{\tiny{\PHplaneTree}}Georgetown University \\
\textsuperscript{\tiny{\PHrosette}}University of Florida \\
\{\emldisplay{xy236@georgetown.edu}{xy236}, \emldisplay{zj153@georgetown.edu}{zj153}, \emldisplay{lb1437@georgetown.edu}{lb1437}, \emldisplay{nathan.schneider@georgetown.edu}{nathan.schneider}\}\texttt{@georgetown.edu} \\
\eml{liu.ying@ufl.edu}
}
\begin{document}
\maketitle
\begin{abstract}
%With Universal Dependencies (UD) treebanks becoming increasingly valuable for NLP and linguistic research, t
CHILDES is a widely used resource of transcribed child and child-directed speech. This paper introduces UD-English-CHILDES, the first officially released Universal Dependencies (UD) treebank. It is derived from previously dependency-annotated CHILDES data, which we harmonize to follow unified annotation principles. The gold-standard trees encompass utterances sampled from 11~children and their caregivers, totaling over 48K sentences (236K tokens). We validate these gold-standard annotations under the UD~v2 framework and provide an additional 1M~silver-standard sentences, offering a consistent resource for computational and linguistic research.
% consolidate existing resources, updating preexisting gold trees to the current UDv2 standard
% gold + silver
\end{abstract}

% \nss{It would be good to have an interesting utterance's tree as a figure on the top right of the first page. Maybe an utterance with a reparandum or other genre-specific characteristics.}

\section{Introduction}
The Child Language Data Exchange System (CHILDES) \citep{macwhinney2000childes} has long been a key resource for research in language acquisition, computational modeling of child language, and the evaluation of Natural Language Processing (NLP) tools. However, many analyses rely on different grammatical assumptions \citep[e.g.,][]{pearl2013syntactic, szubert2024cross, liu-prudhommeaux-2021-dependency, gretz2015parsing, sagae-etal-2007-high}, and therefore adopt divergent annotation frameworks or standards. While most existing annotations use syntactic dependencies—in part due to the relative simplicity of annotation and parsing and the growing adoption of the Universal Dependencies (UD) framework \citep{nivre2016universal, nivre2020universal}—annotation practices remain inconsistent across datasets. This is largely due to the lack of a unified guideline for annotating children’s speech, which presents unique challenges not fully addressed by existing UD documentation.

As UD treebanks have become valuable resources in both NLP \citep[e.g.,][]{jumelet2025multiblimp, opitz2024natural} and language acquisition research \citep[e.g.,][]{clark-etal-2023-cross, hahn2020universals}, there have been increasing efforts to parse CHILDES data using tools such as \texttt{stanza} \citep{liu2024morphosyntactic}. However, the resulting annotation quality is often inconsistent and cannot be guaranteed. In this paper, we compile, harmonize, and manually correct major UD-style annotations of CHILDES data into a consistent, unified UD format, resulting in a gold-standard treebank of 48K sentences and 236K tokens (including, e.g., the tree in \cref{fig:example}). In addition, we construct a larger silver-standard treebank of 1M sentences and 6M tokens produced by \texttt{stanza} \footnote{\texttt{stanza 1.9.2 (combined model)}} and report parser accuracy estimates. We publicly release both datasets\longversion{ to support broader usage in computational and theoretical research}.\footnote{Official gold UD release: \url{https://github.com/UniversalDependencies/UD_English-CHILDES}
Note: Due to a postprocessing error, the gold UD release from the \texttt{main} branch is missing approximately 10K sentences. For complete access to the data, please use the \texttt{dev} branch. The \texttt{main} branch will be updated in the next official release scheduled for November 2025.\\ Silver release: \url{https://github.com/xiulinyang/UD-CHILDES}.}

\begin{figure}[t]
    \centering
    \begin{dependency}
      \begin{deptext}[column sep=2em]
        And \& a \& a \& green \& one \& . \\
      \end{deptext}
      \depedge[edge unit distance=1.7ex]{5}{1}{cc}
      \depedge[edge unit distance=4ex]{3}{2}{reparandum}
      \depedge[edge unit distance=2.5ex]{5}{3}{det}
      \depedge{5}{4}{amod}
      \depedge{5}{6}{punct}
    \end{dependency}
    \caption{UD tree for a child utterance from Lily (Providence corpus, sentID=16916280)}
    \label{fig:example}
\end{figure}

  % \end{subfigure}
  % \hfill
  % \vspace{0.2cm}
  % \begin{subfigure}[t]{0.5\textwidth}
  %   \centering
  %   \begin{dependency}
  %     \begin{deptext}
  %       And \& a  \& purple \& flamingo \& .\\
  %     \end{deptext}
  %     \depedge{4}{1}{cc}
  %     \depedge{4}{2}{det}
  %     \depedge{4}{3}{amod}
  %     \depedge{4}{5}{punct}
  %   \end{dependency}
  %   \caption{UD tree for a child-directed utterance (Lily Providence, sentID=16916265).}
  %   \end{subfigure}
    % \caption{Examples of a child's utterance and child-directed speech respectively\nss{the two utterances are very similar. choose an adult utterance that is a bit different? or omit (b)?}\nss{make sure to reference the figure from the text}}
%     \label{fig:example}
% \end{figure}

\begin{table*}[ht]
\small
\setlength{\tabcolsep}{4.5pt}
\centering
%\resizebox{\textwidth}{!}{
%\begin{threeparttable}

\begin{tabular}{p{2cm}p{1.5cm}p{1cm}llp{0.9cm}rrrr}
\toprule
\textbf{Corpus} & \textbf{Children} & \textbf{Speakers} & \textbf{Trees} &\textbf{UPOS} & \textbf{Feats} & 
\multicolumn{2}{c}{\textbf{Utterances}} & \multicolumn{2}{c}{\textbf{Tokens}} \\
\cmidrule(lr){7-8} \cmidrule(lr){9-10}
& & & & & & \textbf{Gold} & \textbf{Silver} & \textbf{Gold} & \textbf{Silver} \\
\midrule
S+24  & Adam & Adults & Gold &  Gold & Convrtd & 17,233 & 0 & 91,114 & 0 \\
LP21  & Eve & All & Gold & Silver & Silver & 2,207 & 0 & 8,497 & 0 \\
LP23  & 10 Children & All & Gold &  Silver & Silver & 34,530 & 0 & 168,284 & 0 \\
\midrule
\textbf{UD-English-CHILDES} & 11 Children & All & Gold/Silver & Gold/Silver & N/A & \textbf{48,183} & \textbf{1,197,471} & \textbf{236,941} & \textbf{6,892,314} \\
\bottomrule
\end{tabular}
\caption{Overview of CHILDES-based UD treebanks compiled in this paper and our newly-released UD-English-CHILDES treebank. Source corpus labels (S+24, LP21, LP23) are defined in \cref{sec:annos}. Note that there is overlap in the Adam data: S+24 figures are counts from the original dataset; for our version, these were filtered to avoid duplicates and merged with corresponding LP23 utterances. The heading \textbf{Gold} refers to the subset of utterances for which trees and UPOS have been manually corrected according to the UD v2 framework; \textbf{Silver} refers to the subset with fully automatic annotations from \texttt{stanza}.}
\label{overview}
%\end{threeparttable}
%}
\end{table*}

\begin{table*}[ht]
\centering
\small
\resizebox{\textwidth}{!}{
\begin{tabular}{@{}lllrrrrr@{}}
\toprule
\textbf{Child} & \textbf{Corpus} &\textbf{Child age range}    & \multicolumn{1}{c}{\textbf{Gold sents}} & \textbf{Gold toks} & \textbf{Silver sents} & \textbf{Silver toks} \\ \midrule    
Laura & Braunwald \citep{braunwald1971mother}   &   1;3-7;0 (1;3-7;0)    & 4,622                       & 21,079                   & 41,862                         & 205,427                   \\
Adam &Brown  \citep{brown1973first}    &  1;6-5;2 (1;6-5;2)       & 16,736                       & 84,643                   & 93,315                        & 452,348                  \\
Eve & Brown & 1;6-5;1 (1;6-5;2) & 2,207& 8,497 & 108,044 & 532,319\\
Abe & Kuczaj \citep{kuczaj1977acquisition}    &   2;4-5;0 (2;4-5;0)       & 4,167                       & 22,437                   & 38,630                         & 230,489                   \\
Sarah & Brown     &   1;6-5;2 (1;6-5;2)       & 5,347                       & 23,233                   & 104,926                        & 517,654                   \\
Lily & Providence \citep{demuth2006word}   &   0;11-4;0 (0;11-4;0)    & 1,499                       & 6,337                    & 79,573                         & 422,245                   \\
Naima & Providence    &  1;3-3;11 (0;11-4;0)    & 2,534                       & 14,360                   & 236,350                        & 1,422,543                 \\
Violet & Providence   &  0;11-4;0 (0;11-4;0)   & 721                         & 1,857                    & 32,801                         & 164,975                   \\

Thomas & Thomas \citep{lieven2009two}   &    2;0-4;11 (2;0-4;11)   & 4,240                       & 20,333                   & 313,550                        & 2,039,132                 \\
Emma & Weist \citep{weist2008autobiographical}      &   2;2-4;10 (2;1-5;0)     & 2,423                       & 13,730                   & 74,825                         & 474,460                   \\
Roman & Weist      &    2;2-4;9 (2;1-5;0)    & 3,653                       & 20,557                   & 73,595                         & 467,633                   \\

% \midrule
% \textbf{Overall}    & &  & \textbf{48,149}             & \textbf{236,941}         & \textbf{1,197,543}             & \textbf{6,929,436}        \\
\bottomrule
\end{tabular}}
\caption{Detailed statistics for each child, including counts of gold and silver annotations and their corresponding age ranges in months. Ages in the silver corpus are shown in parentheses. For source corpus URLs see \Cref{sources}.}

% \nss{maybe Zoey prefers otherwise but I am used to seeing ages in the format YEAR;MONTHS, e.g. 1;6 for 18 months}\nss{what is the sort order for rows? I think it should be either alphabetical by child or by corpus}
\label{table:lp23}
\end{table*}

\section{Related Work}
\subsection{CHILDES Corpora}
CHILDES is a collection of child–adult conversations recorded in naturalistic or laboratory settings. It has played a central role in both language acquisition research and the development of NLP tools. In addition to specialized corpora—such as clinical datasets \citep{gillam2004test}, naturalistic family interactions \citep{gleason1980acquisition}, and controlled laboratory studies \citep{newman2016input}—CHILDES supports a wide range of approaches to developmental linguistics. Many of its corpora inform foundational theories of language acquisition, particularly the poverty of the stimulus hypothesis \citep{chomsky1976reflections}.
Researchers frequently use child-directed speech from CHILDES to quantify the distribution of linguistic structures that are central to these theories, such as wanna contraction \citep{getz2019acquiring}, anaphoric one \citep{foraker2009indirect, pearl2011far}, auxiliary fronting \citep{perfors2011learnability}, and syntactic islands \citep{pearl2011far}. It has also been used in computational models of language acquisition \citep[e.g.,][]{abend-17}.

CHILDES has also emerged as a valuable resource for NLP tool benchmarking and language model pretraining. Following the work of \citet{huang2016evaluation}, studies such as \citet{liu-prudhommeaux-2023-data} have highlighted the challenges faced by UD parsers when applied to child-directed speech, showing substantial performance gaps compared to adult data. CHILDES also supports recent research on pretraining dynamics \citep{feng-etal-2024-child} and the development of efficient language models, including in initiatives like the BabyLM Challenge \citep{choshen2024call, charpentier2025babylm}.

\begin{figure*}[t]
\tiny
\centering
\begin{lstlisting}[language=,basicstyle=\ttfamily\small,tabsize=6,frame=single]
# sent_id = 22497 (normalized sentence ID across corpora; used to avoid 
                 collisions since some corpora share identical sentence IDs)
# original_sent_id = 946255 (original sentence ID from the corpus, as assigned
                            in childsr)
# childes_toks = who's that (original token string from childsr)
# child_name = Abe
# corpus_name = Kuczaj
# gold_annotation = True
# speaker_age = 43.72369042485472 (child's age in months)
# speaker_gender = male (child's gender)
# speaker_role = Father (speaker role in conversation)
# type = question (sentence type annotation)
# text = Who's that?
1-2	Who's	_	_	_	_	_	_	_	_
1	Who	who	PRON	WP	_	0	root	0:root	_
2	's	be	AUX	VBZ	_	1	cop	1:cop	_
3	that	that	PRON	DT	_	1	nsubj	1:nsubj	SpaceAfter=No
4	?	?	PUNCT	?	_	1	punct	1:punct	_
\end{lstlisting}
\caption{Example of a gold-annotated CoNLL-U sentence from the CHILDES-Providence corpus, with added parenthetical explanations of sentence-level metadata. Enhanced UD (EUD) relations are added deterministically by the script at \url{https://github.com/amir-zeldes/gum/blob/master/_build/utils/eng_enhance.ini}.}
\label{fig:conllu-example}
\end{figure*}

\subsection{Spoken Language Treebanks}

\paragraph{Overview}
The development of UD project has fostered the development of spoken language annotations across a wide variety of languages, such as Beja \citep{kahane2022morph} and Japanese \citep{omura-etal-2023-ud}, as documented in \citet{dobrovoljc2022spoken}. For English, the GUM corpus \citep{zeldes2017gum} incorporates several spoken genres.

\paragraph{CHILDES Dependency Treebanks}
Early dependency parsing research on English CHILDES data %\citep{macwhinney2000childes} 
utilized a custom inventory of grammatical relations \citep[GR;][]{sagae-etal-2004-adding, sagae-etal-2005-automatic}. These gradually evolved to address CHILDES-specific challenges \citep{sagae-etal-2007-high}, and were applied to the entire English CHILDES corpus using a supervised parser \citep{sagae2010morphosyntactic}.

More recently, UD-style annotations have been introduced to CHILDES.  \citet{liu2024morphosyntactic} release an automatically parsed version of the English CHILDES corpus, annotated with UD trees using \texttt{stanza}. \citet{liu-prudhommeaux-2021-dependency} used a semi-automatic method to convert previous GR-based annotations into UD trees, focusing on child-produced speech (ages 18–27 months) from the Eve data within the Brown corpus \citep{brown1973first}. Subsequently, \citet{szubert2024cross} developed gold-standard UD annotations by automatically transforming GR annotations and manually correcting them. Their dataset includes child-directed speech from the Adam data of the Brown corpus and the Hebrew Hagar corpus \citep{berman1990acquiring}, addressing spoken-language-specific phenomena such as repetitions and non-standard vocabulary, as well as a mapping to semantics.

Building upon these efforts, \citet{liu-prudhommeaux-2023-data} significantly expanded UD annotations to cover utterances from 10 children aged 18–66 months (Adam from the Brown corpus as well as 9 children from other corpora), incorporating both child and caregiver speech. Their work tackles complex spoken-language features, including speech repairs and restarts.

Although \citet{liu-prudhommeaux-2021-dependency, liu-prudhommeaux-2023-data} provide manually corrected UD trees, their annotations are inconsistent with the UD~v2 framework, lack Universal Part-of-Speech (UPOS) tags, and have not been independently verified. \citet{szubert2024cross} offer verified data, but they follow the UD~v1 annotation guidelines. To date, there is no official UD release for CHILDES speech data.

\section{Annotations}\label{sec:annos}

\subsection{Data Source \& Statistics}
This work leverages three existing UD treebanks: \citet{szubert2024cross} (henceforth \textbf{S+24}), \citet{liu-prudhommeaux-2021-dependency} (\textbf{LP21}), and \citet{liu-prudhommeaux-2023-data} (\textbf{LP23}), summarized in \cref{overview}. As these treebanks were already annotated, our human annotation efforts focused primarily on correcting errors and harmonizing annotations across corpora. We present post-compilation statistics in \cref{overview,table:lp23}. \Cref{overview} summarizes the full corpus and its source contributions, and \cref{table:lp23} provides per-child statistics. 

In the official UD release, we divide the corpus based on the children’s names and genders. The training and dev splits (90\% and 10\%, respectively) are constructed from the data of Adam, Lily, Naima, Sarah, Roman, Laura, and Abe. The corpora of Eve, Violet, Emma, and Thomas are reserved for the test split. Details are reported in Table~\ref{tab:childes-splits}.

\begin{table*}[h]
\small
\centering
%\resizebox{\linewidth}{!}{
\begin{tabular}{l p{5cm} p{5cm} r}
\toprule
\textbf{Split} & \textbf{Children} & \textbf{Corpus} & \textbf{Gold Sents} \\
\midrule
Train & Adam, Lily, Naima, Sarah, Roman, Laura, Abe & Brown, Providence, Weist, Kuczaj, Braunwald & 34,732 \\
Dev   & Adam, Lily, Naima, Sarah, Roman, Laura, Abe & Brown, Providence, Weist, Kuczaj, Braunwald & 3,860  \\
Test  & Eve, Violet, Emma, Thomas                   & Brown, Providence, Weist, Thomas             & 9,591  \\
\bottomrule
\end{tabular}
%}
\caption{Data splits for the official UD\_English-CHILDES with associated children, corpora, and gold-standard sentence counts.}
\label{tab:childes-splits}
\end{table*}

\subsection{Annotation Pipeline}

Following \citet{liu-prudhommeaux-2023-data}, we collect CHILDES corpora using the R package \texttt{childesr} \citep{sanchez2019childes}.\footnote{\url{https://langcog.github.io/childes-db-website/}} Sentence normalization can be found in the paper. As the data from LP21 and LP23 are only parsed but not tagged yet, sentences with existing dependency annotations are identified and automatically tagged with UPOS using \texttt{stanza} \citep{qi-etal-2020-stanza}, while unannotated sentences are assigned both UPOS and dependency trees. Our current work focuses on correcting previously human-annotated data. To ensure conformity with UD guidelines, we run all processed sentences through the UD validation tool\footnote{\url{https://github.com/UniversalDependencies/tools/blob/master/validate.py}} and manually fix those that fail validation. The correction work is performed by three linguistics graduate students trained in UD annotation. In total, we made approximately 8,000 corrections.

% \nss{elaborate on the major classes of errors encountered and how they were fixed}
Many of the errors stem from mismatches between UPOS tags and dependency labels (as LP21 and LP23 used automatic UPOS tagging).
In addition, we address format issues such as multiword tokens, spacing mismatches (e.g., \texttt{SpaceAfter}), and deprecated dependency relations not supported by current UD guidelines (e.g., \texttt{compound:svc}, \texttt{obl:about\_like}, \texttt{nmod:over\_under}).
The 5 most common linguistic issues were as follows:

\paragraph{\texttt{advmod} tagged as \texttt{ADP}}
This error commonly appears with phrasal verbs such as \textit{get up} and \textit{take over}. The original annotation assigns \texttt{advmod} as the dependency relation to phrasal verbs with POS tag \texttt{ADP}. We revise these to \texttt{compound:prt}, in accordance with the UD treatment of phrasal particles.

\paragraph{Auxiliaries tagged as \texttt{VERB} or \texttt{PART}}
Auxiliaries such as \textit{be} and \textit{have} are frequently misclassified as main verbs or particles. In some cases, lemmas are also mislabeled—most notably, the lemma of contracted forms like \textit{’s} is incorrectly assigned as \textit{'s} rather than the appropriate auxiliary \textit{be}. We correct both the POS and lemma annotations in these cases.

\paragraph{Lexical items tagged as \texttt{PUNCT}}
The \texttt{stanza} parser often mislabels disfluent word fragments in spontaneous speech as punctuation marks (e.g., \textit{OK/INTJ Adam/PROPN ride/VERB dat/\textbf{PUNCT} ./PUNCT}. We reassign these tokens appropriate UPOS labels based on context and speaker intent, often as interjections.

\paragraph{Determiner misrecognition}
Ambiguous or reduced forms of determiners—such as \textit{de} —are frequently misidentified as proper nouns (\texttt{PROPN}). We manually review these cases and reannotate them as \texttt{DET} when appropriate.

\paragraph{Function word heads with dependents}
In previous treebanks, words appearing in functional relations such as \texttt{case}, \texttt{mark}, and \texttt{aux} have been assigned children, which violates UD’s constraint that these words should be leaf nodes. We reassign the erroneous dependents to the appropriate content heads, ensuring the structure conforms to UD’s projectivity and function word constraints.
\subsection{Harmonization}
Each treebank follows its own annotation guidelines, which are largely based on UD but not fully compliant. We performed a series of normalization steps to harmonize them into a consistent format. Our unified format is primarily based on LP23, with several adaptations described below.

\paragraph{Metadata} In our normalized \mbox{CoNLL-U} files, we include the following metadata fields with an example provided in \cref{fig:conllu-example}: \texttt{sent\_id} (normalized sentence IDs); \texttt{original\_sent\_id} (utterance ID retrieved via the \texttt{childesr} R package); 
\texttt{childes\_toks} (tokenized utterance); 
\texttt{corpus\_name} (original corpus name); 
\texttt{gold\_annotation} (indicates whether the sentence is manually annotated); 
\texttt{speaker\_gender}, \texttt{speaker\_role}, and \texttt{speaker\_age} (speaker/child metadata); \texttt{text} (the text aligned with the tree), and 
\texttt{type} (sentence type). %: \textit{declarative}, \textit{question}, \textit{imperative emphatic}, \textit{interruption}, \textit{trail off}, or \textit{quotation next line}). 
\Cref{tab:s_counts} summarizes the distribution of the main sentence types and compares them with those in the UD~2.15 release of GUM \citep{zeldes2017gum}, a multi-genre English corpus.
Notably, questions occur in the CHILDES conversations at a much higher rate---they are nearly half (45\%) as frequent as declarative utterances, as opposed to 9\% in GUM.
\begin{table}[ht]
\centering
\small  
\setlength{\tabcolsep}{4pt}
\begin{tabular}{p{1.2cm}rrrr}
\toprule
\textbf{Type} & \multicolumn{3}{c}{\textbf{CHILDES}} & \multicolumn{1}{c}{\textbf{GUM}} \\
 & \multicolumn{1}{c}{\textbf{CDS}} & \multicolumn{1}{c}{\textbf{CS}} & \multicolumn{1}{c}{\textbf{Overall}} & \multicolumn{1}{c}{\textbf{Overall}}  \\
\cmidrule(l){2-4} \cmidrule(l){5-5}
declarative                 & 16,112 & 15,884 & 31,996 & 7,695 (\texttt{decl})             \\
question                    &  2,882 &  11,413 & 14,295 & 716 (\texttt{q}, \texttt{wh})   \\
imperative emphatic    &   509  &    288 &   797  & 1,326 (\texttt{imp}, \texttt{intj}) \\
others                      &   601  &    494 &1095 &   2,409                         \\
\bottomrule
\end{tabular}
\caption{Sentence type counts in gold CHILDES and GUM corpora. \textbf{Question} includes \textit{question}, \textit{self interruption question}, \textit{trail off question}, and \textit{interruption question}. \textbf{Others} encompasses less frequent categories: \textit{trail off}, \textit{interruption}, \textit{self interruption}, and \textit{quotation next line}.}
\label{tab:s_counts}
\end{table}

% Full distribution:
% Counter({'declarative': 31996,
%          'question': 14241,
%          'imperative_emphatic': 797,
%          'trail off': 695,
%          'interruption': 198,
%          'self interruption': 167,
%          'quotation next line': 35,
%          'self interruption question': 25,
%          'trail off question': 16,
%          'interruption question': 13})
%
% GUM: https://universal.grew.fr/?custom=68031d6f28fc4
% All GUM types: s_type="q"|"wh"|"frag"|"decl"|"sub"|"multiple"|"imp"|"intj"|"inf"|"ger"|"other"

\paragraph{Punctuation}
To bring the transcripts in line with written English conventions, we capitalize the first word of each utterance and infer sentence-final punctuation at the end of each sentence based on the sentence type provided in the metadata.\footnote{The original data transcribes various kinds of prosodic information such as pauses. At present we do not retain this information or attempt to infer corresponding punctuation like commas and parentheses.}

\paragraph{Reparandum}
Each of the three treebanks defines its own subtypes for the \texttt{reparandum} and \texttt{parataxis} relations. For example, S+24 includes labels such as \texttt{parataxis:repeat}  not present in the current UD guidelines. Similarly, LP21 and LP23 annotate \texttt{reparandum} with subtypes such as \texttt{restart} and \texttt{repetition} to mark special utterance features of children's speech. To ensure consistency across treebanks, we move all such subrelation information to the \texttt{MISC} column.

% \nss{explain with examples}

\paragraph{Others}
As S+24 was annotated using the UD guidelines version 1.0, we convert the annotation using UD version 2.0 with a script\footnote{\url{https://github.com/UniversalDependencies/tools/tree/master/v2-conversion}} and manual annotation. For example, we shifted the head-dependent direction of \texttt{flat} in the annotations. 

\begin{table}[]
\small
    \centering
    \resizebox{\linewidth}{!}{\begin{tabular}{cccccc}
    \toprule
    Metrics& Children's speech & Parents' speech & Overall \\ 
    \midrule
    LAS &81.2 &  86.3& 84.2\\
    UAS & 87.2   & 91.0 & 89.5\\
    \bottomrule
    \end{tabular}}
    \caption{LAS and UAS scores for children's speech, parents' speech, and overall performance.}
    \label{tab:uas}
\end{table}

Since S+24 and LP23 overlap in the Adam corpus, we merged the annotations from these two treebanks. 3375 sentences are repetitive in S+24. We removed these sentences from our corpus\footnote{883 sentences from S+24 could not be merged because S+24 and LP21 are using different data sources, and were therefore removed from our treebank as well.}.

To ensure a more linguistically plausible analysis, we also diverged from \citet{liu-prudhommeaux-2023-data} in our treatment of interjections. Instead of annotating utterances consisting solely of interjections (e.g., \textit{Ha ha ha ha}) as \texttt{conj}, we used the \texttt{flat} relation.

\subsection{Silver Data Assessment}
To create silver-standard annotations, we apply \texttt{stanza} to the utterances that were not sampled by the previous treebanks (but were from the same CHILDES datasets, i.e.~conversations involving the 11~children in \cref{table:lp23}). To estimate the quality of these silver annotations, we evaluate the parser’s performance on the gold-standard data. We report Labeled and Unlabeled Attachment Scores (LAS/UAS) in \cref{tab:uas}. The parser achieves an overall LAS of 83.3. Performance is higher on parents’ speech (86.3 LAS) than on children’s speech (81.2 LAS), likely due to the greater syntactic regularity and lower frequency of disfluencies in adult utterances. The overall high-quality data can be more easily verified by human annotators than annotated from scratch. It also provides valuable training data for improving parsers on spoken language.

\section{Conclusion \& Future Work}
In this paper, we present the first harmonized UD treebanks for CHILDES, covering 11 corpora and over 48k sentences from both child-directed and child-produced speech. The three datasets we compiled do not preserve conversational structure, and as a result, the finalized gold-standard treebank lacks coherent dialogue sequencing. Preserving such structure would require additional manual annotation to make sure all sentences are gold. However, since our annotations include the \texttt{original\_sent\_id} field, reconstructing the conversation structure is straightforward. Furthermore, morphological features have not been annotated or independently verified. Future work will focus on further corrections to the silver-standard data and the continued expansion of the treebanks. We welcome collaboration on this ongoing effort.

\section*{Acknowledgments}
We acknowledge Ida Szubert, Omri Abend, Samuel Gibbon, Louis Mahon, Sharon Goldwater, Mark Steedman, and Emily Prud’hommeaux for their contributions to the original UD treebanking efforts. We also thank Brian MacWhinney for helpful discussions and anonymous reviewers for their suggestions.

\bibliography{custom}
% \nocite{Ando2005,andrew2007scalable,rasooli-tetrault-2015}

\appendix
\section{Sources of the Coprora}
\label{sources}
In this work, we include the sources from the following corpora:
\begin{itemize}
    \item \url{https://childes.talkbank.org/access/Eng-NA/Braunwald.html}
    \item \url{
https://childes.talkbank.org/access/Eng-NA/Brown.html}
\item \url{
https://childes.talkbank.org/access/Eng-NA/Kuczaj.html}
\item \url{
https://phon.talkbank.org/access/Eng-NA/Providence.html}
\item \url{
https://childes.talkbank.org/access/Eng-UK/Thomas.html}
\item \url{
https://childes.talkbank.org/access/Eng-NA/Weist.html}
    
\end{itemize}
\end{document}